\documentclass[10pt,twocolumn,letterpaper]{article}

\usepackage{cvpr}              % To produce the CAMERA-READY version
\usepackage{graphicx}
\usepackage{amsmath}
\usepackage{amssymb}
\usepackage{bbding}
\usepackage{booktabs}
\usepackage{multirow}

\usepackage[accsupp]{axessibility} % Improves PDF readability for those with disabilities.

\usepackage[pagebackref,breaklinks,colorlinks]{hyperref}

\usepackage[capitalize]{cleveref}
\crefname{section}{Sec.}{Secs.}
\Crefname{section}{Section}{Sections}
\Crefname{table}{Table}{Tables}
\crefname{table}{Tab.}{Tabs.}

\def\cvprPaperID{4104} % *** Enter the CVPR Paper ID here
\def\confName{CVPR}
\def\confYear{2022}

\begin{document}

\title{Homography Loss for Monocular 3D Object Detection}

% \and

\author{
Jiaqi Gu$^{1,2}$, Bojian Wu$^1$\thanks{Corresponding author: ustcbjwu@gmail.com}, Lubin Fan$^1$, Jianqiang Huang$^1$, Shen Cao$^1$, Zhiyu Xiang$^2$, Xian-Sheng Hua$^1$ \\
$^1$Alibaba Cloud Computing Ltd., $^2$Zhejiang University \\% {\tt\small vadin@zju.edu.cn, ustcbjwu@gmail.com, lubin.flb@alibaba-inc.com, caoshen.cao@alibaba-inc.com},\\
% {\tt\small jianqiang.hjq@alibaba-inc.com, xiangzy@zju.edu.cn, xiansheng.hxs@alibaba-inc.com}
}

\maketitle

%%%%%%%%% ABSTRACT
\begin{abstract}
    Monocular 3D object detection is an essential task in autonomous driving. However, most current methods consider each 3D object in the scene as an independent training sample, while ignoring their inherent geometric relations, thus inevitably resulting in a lack of leveraging spatial constraints. In this paper, we propose a novel method that takes all the objects into consideration and explores their mutual relationships to help better estimate the 3D boxes. Moreover, since 2D detection is more reliable currently, we also investigate how to use the detected 2D boxes as guidance to globally constrain the optimization of the corresponding predicted 3D boxes. To this end, a differentiable loss function, termed as Homography Loss, is proposed to achieve the goal, which exploits both 2D and 3D information, aiming at balancing the positional relationships between different objects by global constraints, so as to obtain more accurately predicted 3D boxes. Thanks to the concise design, our loss function is universal and can be plugged into any mature monocular 3D detector, while significantly boosting the performance over their baseline. Experiments demonstrate that our method yields the best performance (Nov. 2021) compared with the other state-of-the-arts by a large margin on KITTI 3D datasets. %Codes will be released \href{https://github.com/gujiaqivadin/HomographyLoss}{here}.
\end{abstract}

%%%%%%%%% BODY TEXT
\section{Introduction}\label{sec:intro}

Monocular 3D object detection is a fundamental task in computer vision, where the goal is to localize and estimate 3D bounding boxes, parameterized by location, dimension, and orientation, of objects from a single image. It can be applied to various scenes, such as autonomous driving, robotic navigation, etc. However, it is an ill-posed and challenging problem since a single image cannot provide explicit depth information. To acquire such resources, most existing methods resort to LiDAR sensors to obtain accurate depth measurements~\cite{peng2021lidar}, or stereo cameras for stereo depth estimation~\cite{li2019stereo}, but they will increase the cost of practical usages. In comparison, the monocular camera is cost-effective.% and easy to be deployed. Therefore, monocular 3D object detection has gradually attracted more attention.

\begin{figure}[t]
    \centering
    \includegraphics[width=0.98\linewidth]{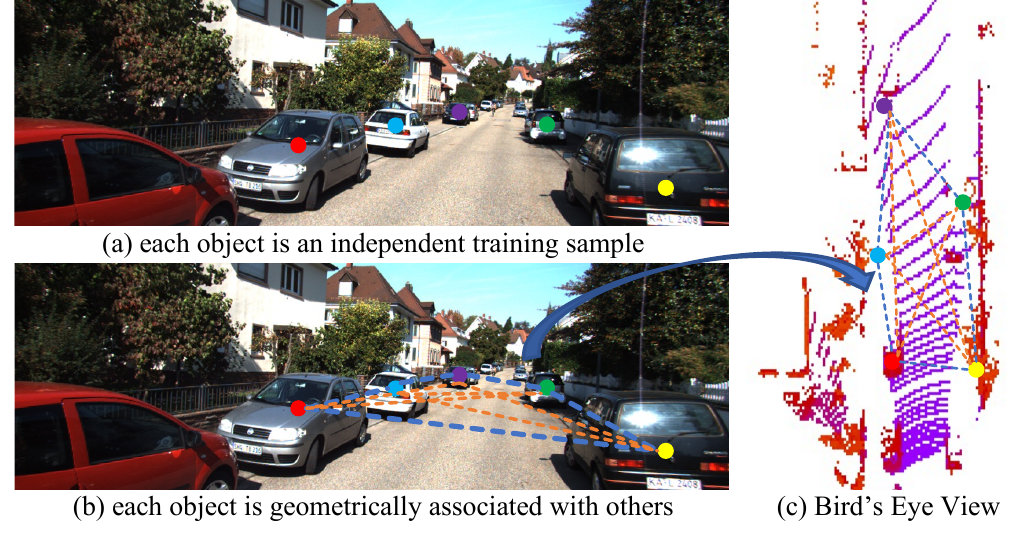}
    \caption{(a) Most of existing methods consider each object as a single training sample, (b) our proposed \textit{homography loss} establishes connections between objects, and applies 2D detection as guidance to help constrain 3D localization in (c) Bird's Eye View.}
    \label{fig:1}
\end{figure}

Most of the existing monocular 3D object detection methods have already achieved remarkable high accuracy with fixed camera settings. However, in their training strategies, each 3D object in the scene is treated as an individual sample without considering the mutual relationships with other neighboring objects, for example, as shown in Fig.~\ref{fig:1}(a). Assuming that, if the predicted 3D box of a single object obviously deviates from its ground truth, without additional constraints, it is usually hard for the network to refine and correct the estimated position of this specific sample. To handle this, apart from the regression loss defined by minimizing the discrepancies between the predicted 3D boxes and the ground truths, many algorithms propose projection loss~\cite{mousavian20173d,li2019stereo,naiden2019shift,li2020rtm3d} to constrain the optimization of 3D boxes with the supervision of corresponding projected 2D ground truth boxes. However, the 3D localization of a single object is still independent of the others. Differently, MonoPair~\cite{chen2020monopair} exploits the object relationships and builds scene graph to enhance the mutual connections of objects during training and inference. They fully leverage the spatial relationships between close-by objects instead of individually focusing on the information-constrained single object. An obvious drawback is that an object can only \textit{locally} connect with its nearest neighbor.

On the other hand, a large percent of approaches are effective for normal objects. In reality, only the foreground objects can be detected easily, because they are fully visible and have rich recognizable features. Therefore, these approaches still struggle to handle the occluded objects or small ones that are far away from the camera, and those objects usually occupy a higher proportion in the scene. Limited improvement is achieved since little information is helpful to solve the problem. A straightforward way to improve the 3D detection is to correct the results by the foreground objects or even the 2D detection results.
% Useful information for those hard cases is naturally scarce, although some recent works aim at solving this problem, while still ending up with limited improvements. A straightforward way to improve the accuracy of detection would be using objects with better detection results (the foreground objects or even 2D detection) to assist in correcting objects with poor detection results.
The most relevant work, MonoFlex~\cite{zhang2021objects}, which leverages the distribution of different objects and proposes a flexible framework to decouple the truncated objects and adaptively combine multiple approaches for 3D detection. However, it is also limited to training the network for each individual sample.% as mentioned before.

Moreover, due to the perspective projection, objects with different depths may block each other in image space. Thus, OFTNet~\cite{roddick2018orthographic} and ImVoxelNet~\cite{rukhovich2021imvoxelnet} propose to regress 3D positions on Bird's Eye View (BEV), since objects on the projected BEV plane do not intersect with each other and can be distinguished.

In general, to be concrete as shown in Fig.~\ref{fig:1}, our core idea is to build the connections between all the objects and globally optimize their 3D positions. Besides, we also associate BEV with image view through inverse projective mapping and apply 2D detection results as guidance to improve the 3D localization in BEV. To achieve the goal, we propose \textit{Homography Loss} to combine 2D and 3D information and globally balance the mutual relationships to obtain more accurate 3D boxes. By doing so, our loss function is able to effectively encode necessary geometric information in both 2D and 3D space, and the network will be enforced to explicitly capture the global geometric relationships between objects which are demonstrated to be helpful for 3D detection. 
Because of the differentiability and interpretability, our loss function can be plugged into any mature monocular 3D detector.
% Because of the concise design, differentiability, and interpretability of our loss function, it can be plugged into any mature monocular 3D detector.
Practically, we take ImVoxelNet~\cite{rukhovich2021imvoxelnet} and MonoFlex~\cite{zhang2021objects} as examples, and integrate the novel homography loss during training phase, experiments demonstrate that our method outperforms the state-of-the-arts by a large margin on KITTI 3D detection benchmark (Nov. 2021). The main contributions can be summarized as follows:

% \begin{itemize}
%     \item We propose a novel loss function, termed as \textit{Homography Loss}, to exploit geometric relationships of all the objects in the scene and globally constrain their mutual locations during training phase. As far as we know, this is the first work that fully leverages the \textit{global} geometric constraints in monocular 3D object detection.
%     \item Since 2D detection is more reliable, our proposed loss function is able to apply 2D results as guidance to correct the 3D predictions, which keeps the consistency of the geometric relationships in both 2D and 3D space.
%     \item The Bird's Eye View contains more information of 3D localization, because the objects on the projected BEV plane will not occlude with each other. Our method leverages the inherent connections between the image plane and BEV plane to enable a better 3D prediction.
% \end{itemize}

\begin{itemize}
    \item We propose a novel loss function, termed as \textit{homography loss}, to exploit geometric relationships of all the objects in the scene and globally constrain their mutual locations, by using the homography between the image view and the Bird's Eye View. At the same time, the geometric consistency in both 2D and 3D space will be well preserved. To the best of our knowledge, this is the first work that fully leverages the \textit{global} geometric constraints in monocular 3D object detection.
    \item The proposed monocular 3D detector based on homography loss achieves the state-of-the-art performance on KITTI 3D detection benchmark, and surpasses the results of all the other monocular 3D detectors, which implies the superiority of our loss.
    \item We apply this loss function to several popular monocular 3D detectors. Without any additional inference cost, the training is more stable and easier to converge, achieving higher accuracy and performance. It can be a plug-and-play module and be adapted to any monocular 3D detector.
\end{itemize}
\section{Related Work}\label{sec:related}

We first review methods on monocular 3D object detection, followed by a brief introduction of geometric constraints that are commonly used during training phase.

% and visual relationship used in image understanding tasks, which are closely related to our proposed method.

\textbf{Monocular 3D object detection} is an ill-posed problem because of lacking depth clues of the monocular 2D image. When compared with stereo images~\cite{li2019stereo} or LiDAR-based methods~\cite{wang2019pseudo,weng2019monocular,qian2020end,ma2019accurate,park2021pseudo,reading2021categorical}, in some earlier works, auxiliary information are necessary for monocular 3D detection to achieve competitive results. These prior knowledge usually includes ground plane assumption~\cite{chen2016monocular}, morphable wireframe model hypothesis~\cite{he2019mono3d++} or 3D CAD model~\cite{chabot2017deep,kundu20183d}, etc.

% \textbf{Monocular 3D object detection} is an ill-posed problem because of lacking depth clues of monocular 2D image, thus compared with stereo images~\cite{li2019stereo} or LiDAR-based methods~\cite{wang2019pseudo,weng2019monocular,qian2020end,ma2019accurate,park2021pseudo,reading2021categorical}, in some earlier works~\cite{murthy2017reconstructing,xiang2017subcategory}, prior knowledge or auxiliary information are necessary for monocular 3D object detection, such as ground plane assumption~\cite{chen2016monocular}, morphable wireframe model hypothesis~\cite{he2019mono3d++} or 3D CAD model~\cite{chabot2017deep,kundu20183d} etc.

%Mono3D~\cite{chen2016monocular} assumes that 3D objects are on the ground plane. Mono3D++~\cite{he2019mono3d++} learns 3D shape hypothesis via morphable wireframe model. Given the shape prior and keypoints detections~\cite{murthy2017reconstructing}, the network estimates the 3D pose and shape of a vehicle. Deep MANTA~\cite{chabot2017deep} proposes a semi-automatic annotation process to generate labels on training images using 3D CAD models. 3D-RCNN~\cite{kundu20183d} makes use of rich shape priors by learning a class-specific low dimensional embedding of shapes from CAD model collections. SubCNN~\cite{xiang2017subcategory} explores how subcategory information can be exploited in CNN-based object detection.

Moreover, some other works only take a single RGB image as input. For example, Deep3DBox~\cite{mousavian20173d} estimates the 3D pose and dimension from the image patch enclosed by a 2D box. Afterwards, the network with a 3D regression head~\cite{liu2019deep,naiden2019shift,fang20193d} is used to predict the 3D box while searching and filtering the proposal whose 2D projection has the threshold overlap with the ground-truth 2D box. MonoGRNet~\cite{qin2019monogrnet} detects and localizes 3D boxes via geometric reasoning in both the observed 2D projection and the unobserved depth dimension. MonoDIS~\cite{simonelli2019disentangling} leverages a novel disentangling transformation for 2D and 3D detection losses. M3D-RPN~\cite{brazil2019m3d} reformulates the monocular 3D detection problem as a standalone 3D region proposal network. Unlike previous methods, which depend on 2D proposals, SMOKE~\cite{liu2020smoke} argues that the 2D detection network is redundant and introduces non-negligible noise in 3D detection. Thus, it predicts a 3D box for each object by combining a single keypoint estimated with regressed 3D variables via a single-stage detector, and similarly, RTM-3D~\cite{li2020rtm3d} predicts nine perspective keypoints of a 3D box in the image space. Specifically, MonoFlex~\cite{zhang2021objects} proposes a flexible framework for monocular 3D object detection that explicitly decouples the truncated objects and adaptively combines multiple approaches for depth estimation.%object depth estimation.

% Starting with the pioneering work of~\cite{wang2019pseudo,weng2019monocular}, these methods~\cite{qian2020end,ma2019accurate,park2021pseudo,reading2021categorical} leverage advances in monocular depth estimation and lift the input image to a pseudo-LiDAR point cloud representation, and then train an end-to-end LiDAR-based 3D detector to achieve impressive performance. 

%The previous methods use off-the-shelf depth estimation to assist 3D detection and gain performance improvement. However, to be simplified, DD3D~\cite{park2021pseudo} proposes an end-to-end, single stage monocular 3D object detector that can benefit from depth pre-training like pseudo-LiDAR methods and enables an effective information transfer between depth estimation and 3D detection, and CaDDN~\cite{reading2021categorical} also tries to integrate the dense depth estimation into monocular 3D detection. 

However, image-based training and inference will introduce non-linear perspective distortion where the scale of objects varies drastically with depth, which makes it hard to accurately predict the relative distance and location of the object of interest. 
To handle this, OFTNet~\cite{roddick2018orthographic} proposes orthographic feature transform by mapping image-based features into an orthographic 3D space that is better aligned with the real-world perception, where the target object will not intersect or occlude with each other and can be intuitively distinguished.
% To handle this, OFTNet~\cite{roddick2018orthographic} proposes orthographic feature transform by mapping image-based features into an orthographic 3D space, which could be better aligned with the real world perception and in which space the target object will not intersect and occlude with each and can be intuitively distinguished.
ImVoxelNet~\cite{rukhovich2021imvoxelnet} projects the obtained image features extracted from the backbone network to a 3D voxel volume and proposes to detect 3D boxes from BEV for the same purpose as just mentioned. 

% 对mono3d做一个小总结，其实要么是在image view回归，要么是直接在3d(bev) view回归，没有将这两个view之间的相互关系联系起来约束，我们的loss做到了这一点

Overall, current methods consider either directly regressing depth or keypoints from image view, or detecting 3D boxes from BEV. As none of the existing methods dig into the inherent connection between the image view and BEV, our proposed method first bridges the gap between them. %We also take all the objects into account and explore their mutual relationships for better refining 3D boxes, instead of treating each object as a single training sample.

% https://towardsdatascience.com/geometric-reasoning-based-cuboid-generation-in-monocular-3d-object-detection-5ee2996270d1

\begin{figure}[t]
    \centering
    \includegraphics[width=0.9\linewidth]{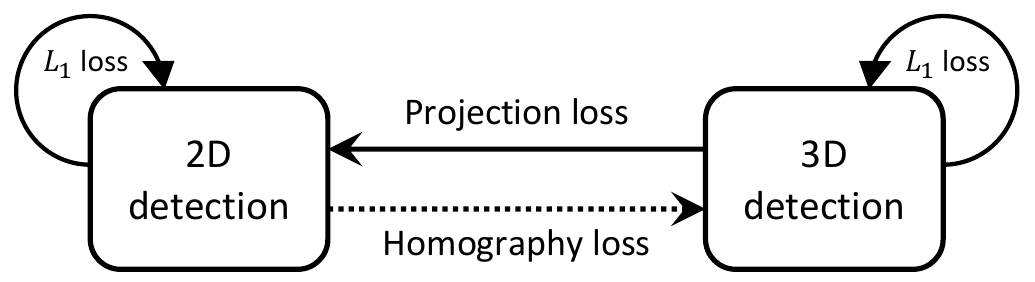}
    \caption{In practice, for 2D/3D detection tasks, the discrepancies between the predicted 2D/3D boxes and the corresponding ground truths can be narrowed down by applying \rm{L1} loss. It means that the predicted 2D/3D boxes would be self-constrained with the corresponding ground truths. Besides, the predicted 3D positions of objects can be projected into 2D image space with camera parameters, and the projected 2D positions will be further compared with their 2D ground truths. That is to say, 3D data can be converted to 2D space via a projection matrix. By analogy, our proposed method builds the correlation from 2D to 3D and uses 2D detection as guidance to supervise the training of 3D localization.}
\label{fig:2}
\end{figure}

\textbf{Geometric constraints in 3D detection}. Most current approaches directly regress 3D spatial information from 2D image without the help of extra 3D priors. Because 2D and 3D space are naturally interrelated via perspective projection, therefore, some recent works attempt to use geometric constraints in the network. Mousavian~\etal~\cite{mousavian20173d} estimates 3D boxes using the geometric relations between 2D edges and 3D corners. Li~\etal~\cite{li2019stereo} solves a coarse 3D box by utilizing the sparse perspective keypoints and 2D box. Naiden~\etal~\cite{naiden2019shift} solves the translation vector of the object center via a closed-form least squares equation. Li~\etal~\cite{li2020rtm3d} utilizes the geometric relationship of 3D and 2D perspectives to recover 3D boxes. Li~\etal~\cite{li2021monocular} reformulates the non-linear optimization in the projective space as a differentiable geometric reasoning module. Note that, the aforementioned methods apply the geometric constraint to individual object. Contrarily, we take the positional relationships of all the objects into account at the same time.

\section{Methods}\label{sec:methods}

\subsection{Motivation}\label{subsec:motivation}

\begin{figure}[t]
    \centering
    \includegraphics[width=0.98\linewidth]{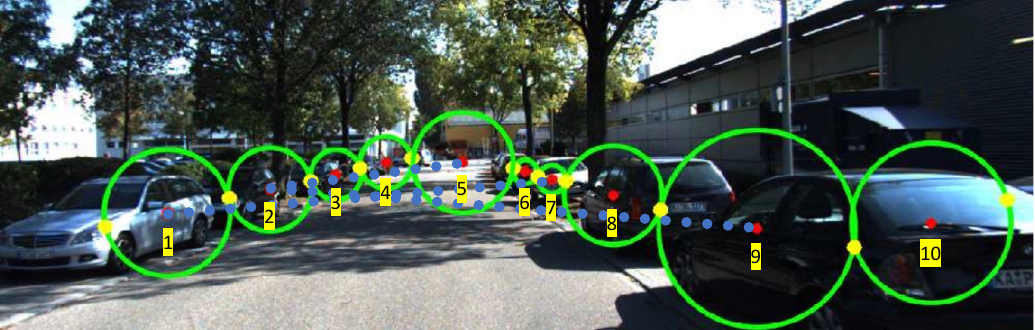}
    % \caption{The pairwise relationship is not enough to encode the whole spatial relationship of objects. As proposed in ~\cite{chen2020monopair}, a single object can only \textit{locally} connect with its nearest neighbor. However, for a single object, the positions of others should \textit{globally} have an impact on the localization of the target object, which mimics the long-range dependency used in the attention mechanism. For example, the location of Car 2 can not only be influenced by Car 1, but also be constrained by Car 5 and 9 as connected with the blue dotted line. (The figure originates from \cite{chen2020monopair})}
    \caption{The location of the target object is globally affected by other objects. Since an individual object can only locally connect with its nearest neighbor, as proposed in ~\cite{chen2020monopair}, the pairwise relationship is not enough to encode the spatial relationship of objects. We take the global affection into account, which is similar to the long-range dependency used in the attention mechanism. For example, the location of Car 2 can not only be influenced by Car 1, but also be constrained by Car 5 and 9 as connected with the blue dotted line. (The figure originates from \cite{chen2020monopair})}
    \label{fig:3}
\end{figure}

We have two key observations: 1) the 2D detection can serve as a guidance to constrain and supervise the training of 3D localization, 2) the position of a single object should be \textit{globally} influenced by the surrounding objects, as detailed in Fig.~\ref{fig:2} and~\ref{fig:3}. To handle those problems, we propose \textit{homography loss} to implement the conversion from 2D image space to 3D BEV space, and simultaneously constrain the globally geometric relationships of all the objects.

\subsection{Revisiting of Homography}\label{subsec:homo}
A homography is a mapping between two planar surfaces which preserves collinearity. The homography matrix $\mathbf{H} \in \mathbb{R}^{3 \times 3}$ between two 2D planes maps $\mathbf{p}_1$ in the plane 1 to $\mathbf{p}_2$ in the plane 2 up to a scale factor $s$. It satisfies:
\begin{equation}
s \mathbf{p}_2 = \mathbf{H} \mathbf{p}_1,
\label{eq:homo}
\end{equation}
where $\mathbf{p}=[x,y,1]^T$ is the homogeneous coordinate of a 2D point in a plane. Since the homography matrix has 8 degrees of freedom, at least 4 corresponding point pairs are necessary for recovering the matrix. Inspired by ImVoxelNet~\cite{rukhovich2021imvoxelnet}, the projections of objects on BEV plane do not intersect with each other and accordingly contain more information about 3D localization, we define the homography matrix between the image plane and BEV plane, in order to implicitly transform coordinates from 2D to 3D space.
More details will be illustrated in Sec.~\ref{subsec:homo_loss}. Then, let us explain why homography is a global geometric constraint. Firstly, all pairs of corresponding points will involve in solving the homography matrix from Eq.~\ref{eq:homo}, and the solution is guaranteed to be globally optimal. In other words, the constraint enforced by arbitrary pair of corresponding points will finally affect the whole optimization process. Thus, homography is a global constraint. Secondly, in projective geometry, a homography is an isomorphism of projective spaces, which correlates a group of points on one plane to the other and preserves geometric properties, e.g., collinearity. So, homography is also a geometric constraint.

\subsection{Homography Loss}\label{subsec:homo_loss}
 
% Many previous monocular 3D detection methods use 2D/3D geometric information to better localize objects. Some methods~\cite{brazil2019m3d,li2020rtm3d} associate 3D and 2D keypoints of a single object with the projection matrix, and formulate it as a nonlinear optimization problem which will be solved by minimizing the reprojection error. Other methods, such as MonoPair~\cite{chen2020monopair}, take the pairwise geometric constraint among adjacent objects into consideration and also propose a post-optimization module to further refine the predicted position. The major issues have been discussed in Sec.~\ref{subsec:motivation}. In addition to this, post optimization is not learning-based and cannot be plugged into an end-to-end detector. How to apply strong geometric constraints globally among all objects and make it differentiable is still a tough problem.

Inspired by those observations, we propose a global loss function, termed as \textit{homography loss}, aiming to establish the geometric connections among all the objects by leveraging the homography matrix. Assuming that we already have a monocular 3D object detector that could predict 3D boxes under the supervision of the ground truths, in addition to the regular classification and regression loss in the common pipelines, our homography loss penalizes the wrong relationship among all the predicted boxes and refines the final locations. The major steps are listed as follows.

\begin{figure}[t]
    \centering
    \includegraphics[width=0.98\linewidth]{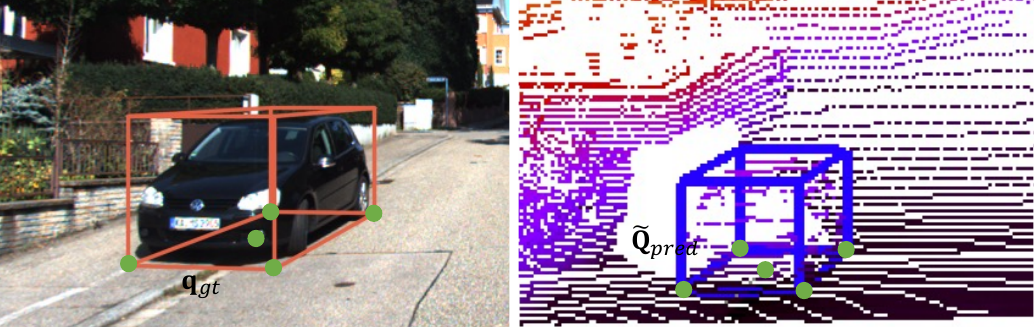}
    \caption{2D and 3D candidate points of a single object.}
    \label{fig:4}
\end{figure}

% 这里给一个box的图例，表明底面的五个点
\textbf{Candidate Points Modeling.} Suppose we have the predicted boxes $\rm{box_{pred}}$ obtained from the arbitrary 3D detector and the corresponding ground truth boxes $\rm{box_{gt}}$.
% and each box can be denoted as $\mathbf{b} = (x, y, z, w, l, h, \theta)$, where $(x, y, z)$ is the 3D location in camera coordinates, $(w, l, h)$ are 3D object dimensions, and $\theta$ is rotation angle around y-axis.
As mentioned in Sec.~\ref{subsec:homo}, we opt to use the homography matrix to describe the projection relationship between the image plane and the BEV plane. For a single object, as demonstrated in Fig.~\ref{fig:4}, we pick up five bottom points $\mathbf{Q}_{pred} = [x_{pred}, y_{pred}, z_{pred}]^T$ of $\rm{box_{pred}}$ as representatives, including one bottom center point and four bottom corner points. We also assume that all the objects are always on the flat ground, the bottom points on the BEV plane can thus be simplified as $\tilde{\mathbf{Q}}_{pred} = [x_{pred}, y_{pred}]^T$. Similarly, we have $\mathbf{Q}_{gt} = [x_{gt}, y_{gt}, z_{gt}]^T$ obtained from $\rm{box}_{gt}$. After the camera projection, the ground truth 3D box will be transformed into the image space, which is defined by:
\begin{equation}
\mathbf{q} = \mathbf{K} \left[ \mathbf{R}|\mathbf{t} \right] \mathbf{Q},
\label{eq:proj}
\end{equation}
where $\mathbf{K}$ is the intrinsic matrix and $[\mathbf{R}|\mathbf{t}]$ are the extrinsic matrices, and $\mathbf{q} = [u, v]^T$ represents the projected pixel on the image plane, which is suitable for both $\rm{box}_{pred}$ and $\rm{box}_{gt}$. Therefore, if there exist $N$ objects, we can get $5N$ pairs of candidate points $\mathbf{q}_{pred}$, $\tilde{\mathbf{Q}}_{pred}$ for $\rm{box}_{pred}$ and $\mathbf{q}_{gt}$, $\tilde{\mathbf{Q}}_{gt}$ for $\rm{box}_{gt}$, respectively, which are prepared for calculating the homography matrix.
    
% 加入一张场景中各物体底面中心相连的图，包括Imagev和bev。

\textbf{Calculating Homography.} To implicitly constrain relative positions of each object, without loss of generality, we select $\mathbf{q}_{gt}$ and $\tilde{\mathbf{Q}}_{pred}$. Specifically, we use the ground truth coordinates $\mathbf{q}_{gt}$ in 2D image view as guidance, to correct the final positions $\tilde{\mathbf{Q}}_{pred}$ in 3D space. The formulation is defined, up to a scale factor (omitted here) with homogeneous coordinates, as follows,
\begin{equation}
    \tilde{\mathbf{Q}}_{pred} = \mathbf{H} \mathbf{q}_{gt}, 
    \begin{pmatrix} x_{pred} \\ y_{pred} \\ 1 \end{pmatrix} = \mathbf{H} \begin{pmatrix} u_{gt} \\ v_{gt} \\ 1 \end{pmatrix}.
\label{eq:calc_H}
\end{equation}

Here, $\mathbf{H}$ stores the mutual relationships of all the objects by mapping between two views. We use singular value decomposition (SVD) to calculate the homography matrix $\mathbf{H}$ as it can be easily implemented in PyTorch~\cite{paszke2019pytorch}.

% Note that, in practice, at the very beginning of training, $\tilde{\mathbf{Q}}_{pred}$ may deviate a lot from the ground truths, thus, the solution of Eq.~\ref{eq:calc_H} is estimated and approximated. 
In practice, the homography matrix in Eq.~\ref{eq:calc_H} is estimated since $\tilde{\mathbf{Q}}_{pred}$ may deviate a lot from the ground truth at the very beginning of training.
We denote it as $\hat{\mathbf{H}}$, and represent $\tilde{\mathbf{Q}}_{homo}$ = $\hat{\mathbf{H}} \mathbf{q}_{gt}$. As the training progresses, the estimated value $\tilde{\mathbf{Q}}_{homo}$ will approach $\tilde{\mathbf{Q}}_{pred}$ and $\tilde{\mathbf{Q}}_{gt}$.

\begin{table*}[t!]
    \begin{center}
    \caption{3D object detection performance of Car category on KITTI \textit{test} set. The best scores are marked in bold (compared with monocular 3D detection methods that do not use extra data). 'Extra Data' lists the required extra information for each method, including \textbf{Depth} pretrained from a much larger dataset, \textbf{Temporal} information from multi frames, \textbf{LiDAR} with point cloud information and \textbf{Shape} of extra labelled 3D instance keypoints. All runtime values are collected from KITTI benchmark as well as the official paper and code.}
    \label{table:kittitest}

    \scalebox{0.75}{
    \begin{tabular}{c|c|c|c|c|c|c|c|c}
    \toprule
    \multirow{2}{*}{Method} & \multirow{2}{*}{Extra Data}  & \multicolumn{3}{|c|}{$\textup{AP}_{3D|R_{40}}$}  &\multicolumn{3}{|c|}{$\textup{AP}_{BEV|R_{40}}$} & \multirow{2}{*}{Time(s)} \\ 
    \cmidrule{3-8}
    &  &  Easy & Moderate & Hard & Easy & Moderate & Hard  \\
    \midrule
    % point
    Mono-PLiDAR \cite{weng2019monocular}  & Depth & 10.76 & 7.50 & 6.10 & 21.27 & 13.92 & 11.25 & 0.10 \\ 
    PatchNet \cite{Ma2020RethinkingPR} & Depth  & 15.68 & 11.12 & 10.17 & 22.97 & 16.86 & 14.97 & 0.40 \\
    D4LCN \cite{Ding2020LearningDC} & Depth  & 16.65 & 11.72 & 9.51 & 22.51 & 16.03 & 12.55 & 0.20 \\
    MonoRUn \cite{Chen2021MonoRUnM3} & Depth  & 19.65 & 12.30 & 10.58 & 27.94 & 17.34 & 15.24 & 0.07  \\
    Kinematic3D \cite{Brazil2020Kinematic3O} & Temporal  & 19.07 & 12.72 & 9.17 & 26.69 & 17.52 & 13.10 & 0.12 \\
    DDMP-3D \cite{Wang_2021_CVPR}  & Depth & 19.71 & 12.78 & 9.80 & 28.08 & 17.89 & 13.44 & 0.18 \\
    Aug3D-RPN \cite{he2021aug3drpn} & Depth  & 17.82 & 12.99 & 9.78 & 26.00 & 17.89 & 14.18 & 0.08 \\
    DFR-Net \cite{Zou_2021_ICCV} & Depth  & 19.40 & 13.63 & 10.35 & 28.17 & 19.17 & 14.84 & 0.18 \\
    CaDDN \cite{reading2021categorical} & LiDAR & 19.17 & 13.41 & 11.46 & 27.94 & 18.91 & 17.19 & 0.63 \\
    MonoEF \cite{Zhou_2021_CVPR} & Depth  & 21.29 & 13.87 & 11.71 & 29.03 & 19.70 & 17.26 & 0.03 \\
    Autoshape \cite{Liu_2021_ICCV} & Shape  & 22.47 & 14.17 & 11.36 & 30.06 & 20.08 & 15.59 & 0.04 \\
    % DD3D \cite{park2021pseudo} & Depth+LiDAR & 23.22 & 16.34 & 14.20 & 30.98 & 22.56 & 20.03 & N/A \\

    \midrule
    % OFTNet \cite{x} & -  & 1.61 & 1.32 & 1.00 & 7.16 & 5.69 & 4.61 & n/a \\
    M3D-RPN \cite{brazil2019m3d} & -  & 14.76 & 9.71 & 7.42 & 21.02 & 13.67 & 10.23 & 0.16 \\
    SMOKE \cite{liu2020smoke} & -  & 14.03 & 9.76 & 7.84 & 20.83 & 14.49 & 12.75 & 0.03  \\
    MonoPair \cite{chen2020monopair} & -  & 13.04 & 9.99 & 8.65 & 19.28 & 14.83 & 12.89 & 0.06  \\
    RTM3D \cite{li2020rtm3d} & -  & 14.41 & 10.34 & 8.77 & 19.17 & 14.20 & 11.99 & 0.05  \\
    PGD-FCOS3D \cite{wang2021probabilistic} & -  & 19.05 & 11.76 & 9.39 & 26.89 & 16.51 & 13.49 & 0.03 \\
    M3DSSD \cite{Luo_2021_CVPR} & -  & 17.51 & 11.46 & 8.98 & 24.15 & 15.93 & 12.11 & 0.16 \\
    MonoDLE \cite{Ma_2021_CVPR}  & -  & 17.23 & 12.26 & 10.29 & 24.79 & 18.89 & 16.00 &  0.04 \\
    MonoRCNN \cite{Shi_2021_ICCV}  & -  & 18.36 & 12.65 & 10.03 & 25.48 & 18.11 & 14.10 & 0.07 \\
    % Ground-Aware \cite{Liu2021GroundAwareM3}  & -  & 21.65 & 13.25 & 9.91 & \textbf{29.81} & 17.98 & 13.08 & 0.05 \\
    
    \midrule
    ImVoxelNet \cite{rukhovich2021imvoxelnet}  & -  & 17.15 & 10.97 & 9.15 & 25.19 & 16.37 & 13.58 & 0.20 \\
    ImVoxelNet(+homo)  & -  & 20.10 & 12.99 & 10.50 & 29.18 & 19.25 & 16.21 & 0.20 \\
    MonoFlex \cite{zhang2021objects}  & -  & 19.94 & 13.89 & 12.07 & 28.23 & 19.75 & 16.89 & 0.03 \\
    MonoFlex(+homo)  & -  & \textbf{21.75} & \textbf{14.94} & \textbf{13.07} & \textbf{29.60} & \textbf{20.68} & \textbf{17.81} & \textbf{0.03} \\
    \bottomrule
    \end{tabular}}
    % \vspace{-0.7cm} 
    \end{center}
\end{table*}

% 是否需要介绍svd求解Homography的基本原理

\textbf{Loss Function.} The homography matrix $\hat{\mathbf{H}}$ implicitly contains the correspondences between two different views and the relative positions of all the objects. Previously, 3D detection is treated as an independent task for each object, which is constrained by regression loss, such as $ L_{reg} = \rm{L1} \left( \tilde{\mathbf{Q}}_{gt} - \tilde{\mathbf{Q}}_{pred} \right) $. Here, we propose a novel loss function, named as \textit{homography loss}, to optimize the locations with strong spatial constraints. The homography loss is defined as follows,
\begin{equation}
\begin{aligned}
    L_{homo} &= \rm{SmoothL1} \left( \tilde{\mathbf{Q}}_{gt} - \tilde{\mathbf{Q}}_{homo} \right) \\
    ~ &= \rm{SmoothL1} \left( \tilde{\mathbf{Q}}_{gt} - \hat{\mathbf{H}} \mathbf{q}_{gt} \right).
\end{aligned}
\label{eq:loss_homo}
\end{equation}

Different from the regression loss, calculating the homography matrix $\hat{\mathbf{H}}$ will take all pairs of corresponding points into consideration. It is therefore a global loss for geometric constraint, which is used to guide the prediction of 3D positions from the ground truth 2D localization. On the other hand, by optimizing Eq.~\ref{eq:loss_homo}, $\hat{\mathbf{H}}$ is also enforced to be closer to the ground truth homography matrix. Another advantage of homography loss is that it is differentiable. It can be a plug-and-play module for any monocular 3D object detector, and servers as  a strong spatial constraint for 3D localization of objects.

\subsection{Case Study}

As our novel homography loss can be plugged into any 3D object detector, we take the state-of-the-art detectors, ImVoxelNet\cite{rukhovich2021imvoxelnet} and MonoFlex\cite{zhang2021objects}, as examples, and illustrate how to seamlessly integrate our loss function into the network. As the main algorithm has been explained in Sec.~\ref{subsec:homo_loss}, more details of the selection of predicted boxes and training strategies are presented here.

\textbf{Anchor based method.} ImVoxelNet\cite{rukhovich2021imvoxelnet} is a one-stage anchor-based monocular 3D detector, which transforms 2D image features into 3D space and regresses the positions of objects in BEV like LiDAR-based 3D detectors. Anchors with IoU $>$ 0.6 will be considered as positives for training and each ground truth object will be assigned by several positive anchors that are served as potential proposals.

To calculate homography, we need to specify one-to-one matching point pairs for the predicted boxes and the ground truth boxes. Therefore, we choose the one with the highest classification score from positive proposals as a representative, which also keeps the consistency between classification and regression. As anchor-based detectors always produce stable proposals during training, we add the homography loss at the beginning of training and train the network from scratch. The loss function defined below consists of four parts, i.e., location loss $L_{loc}$, focal loss for classification $L_{cls}$, cross-entropy loss for direction $L_{dir}$ , and additional homography loss $L_{homo}$:
\begin{equation}
\resizebox{.88\hsize}{!}{$
L = \frac{1}{N_{pos}} \left( \lambda_{cls} L_{cls} + \lambda_{loc} L_{loc} + \lambda_{dir} L_{dir} + \lambda_{homo} L_{homo} \right)$},
\end{equation}
where $N_{pos}$ is the number of positive anchors, $\lambda_{cls}= 1.0, \lambda_{loc} = 2.0, \lambda_{dir} = 0.2, \lambda_{homo} = 0.2$. Note that, apart from $L_{homo}$, other loss terms and balancing weights are all adopted from \cite{rukhovich2021imvoxelnet}.

\textbf{Anchor-free based method.} MonoFlex~\cite{zhang2021objects} is a one-stage monocular 3D detector based on CenterNet~\cite{zhou2019objects}, which predicts projected 3D center, box (including depth, dimension, and orientation), and keypoints in different heads. As it is an anchor-free detector, the location of the representative box is automatically assigned as the 3D projected center in the heatmap head without selection. And the depth is regressed in the final head. The main difference is the training policy.

As 3D projected center and depth can define the coordinates in the image view and Bird's Eye View, these two components are the main contributors for homography loss. But the depth head is very unstable at the beginning of training, and the locations in the Bird's Eye View is also of low confidence, making the homography matrix distorted. Therefore, two strategies are proposed to solve the problem. Firstly, we make a delay by adding our homography loss after 40 epochs when the depth head is consistent and reliable. Secondly, we replicate the predicted boxes by using one of the components (3D projected center and depth), while replacing the other one with its ground truth values. Therefore, homography loss can be replicated three times and ensembled together. The main loss function can be described as a combination of classification loss for heatmap $L_{hm}$, regression loss for box size and rotation $L_{box}$, regression loss for keypoints of 3D boxes $L_{kp}$, and additional homography loss $L_{homo}$:
\begin{equation}
\resizebox{.88\hsize}{!}{$
L = \frac{1}{N_{pos}}(\lambda_{hm}L_{hm} + \lambda_{box}L_{box} + \lambda_{kp}L_{kp} + \lambda_{homo}L_{homo})$},
\end{equation}
where $N_{pos}$ is the number of positive predictions, $\lambda_{hm}= 1.0, \lambda_{box} = 1.0, \lambda_{kp} = 1.0, \lambda_{homo} = 0.2$.

\section{Experiments}\label{sec:exp}

\subsection{Setup}\label{subsec:setup}

\begin{table}[t]
    \caption{3D object detection performance of Car category on KITTI \textit{validation} set.}
    \begin{center}
    \scalebox{0.7}{
    \begin{tabular}{c|c|c|c|c|c|c}
    \toprule
    \multirow{2}{*}{Method} & \multicolumn{3}{|c|}{$\textup{AP}_{3D|R_{40}}$}  &\multicolumn{3}{|c}{$\textup{AP}_{BEV|R_{40}}$}  \\ 
    \cmidrule{2-7}
        & Easy & Moderate & Hard & Easy & Moderate & Hard  \\
    \midrule
    M3D-RPN \cite{brazil2019m3d} & 14.53 & 11.07 & 8.65 & 20.85 & 15.62 & 11.88  \\
    MonoPair \cite{chen2020monopair} & 16.28 & 12.30 & 10.42 & 24.12 & 18.17 & 15.76  \\
    MonoRCNN \cite{Shi_2021_ICCV} & 16.61 & 13.19 & 10.65 & 25.29 & 19.22 & 15.30  \\
    MonoDLE \cite{Ma_2021_CVPR} & 17.45 & 13.66 & 11.68 & 24.97 & 19.33 & 17.01  \\
    \midrule
    ImVoxelNet(+homo) & 21.44 & 14.88 & 12.08 & 29.85 & 21.17 & 17.77  \\
    MonoFlex(+homo) & \textbf{23.04} & \textbf{16.89} & \textbf{14.90} & \textbf{31.04} & \textbf{22.99} & \textbf{19.84}  \\
    \bottomrule
    \end{tabular}}
    \end{center}
    \label{table:kittival}
\end{table}

\begin{table}[t]
    \caption{3D object detection performance of Pedestrian and Cyclist on KITTI \textit{test} set.}
    \begin{center}
    \scalebox{0.7}{
    \begin{tabular}{c|c|c|c|c|c|c}
    \toprule
    \multirow{2}{*}{Method} & \multicolumn{3}{|c|}{Pedestrian $\textup{AP}_{3D|R_{40}}$}  &\multicolumn{3}{|c}{Cyclist $\textup{AP}_{3D|R_{40}}$}  \\ 
    \cmidrule{2-7}
     & Easy & Moderate & Hard & Easy & Moderate & Hard  \\
    \midrule
    PGD-FCOS3D \cite{wang2021probabilistic} & 2.28 & 1.49 & 1.38 & 2.81 & 1.38 & 1.20  \\
    MonoEF \cite{Zhou_2021_CVPR} & 4.27 & 2.79 & 2.21 & 1.80 & 0.92 & 0.71  \\
    D4LCN \cite{Ding2020LearningDC} & 4.55 & 3.42 & 2.83 & 2.45 & 1.67 & 1.36  \\
    M3D-RPN \cite{brazil2019m3d} & 4.92 & 3.48 & 2.94 & 0.94 & 0.65 & 0.47  \\
    DDMP-3D \cite{Wang_2021_CVPR} & 4.93 & 3.55 & 3.01 & 4.18 & 2.50 & 2.32  \\
    DFR-Net \cite{Zou_2021_ICCV} & 6.09 & 3.62 & 3.39 & \textbf{5.69} & \textbf{3.58} & \textbf{3.10}  \\
    M3DSSD \cite{Luo_2021_CVPR} & 5.16 & 3.87 & 3.08 & 2.10 & 1.51 & 1.58  \\
    Aug3D-RPN \cite{he2021aug3drpn} & 6.01 & 4.71 & 3.87 & 4.36 & 2.43 & 2.55  \\
    MonoFlex \cite{zhang2021objects} & 9.43 & 6.31 & 5.26 & 4.17 & 2.35 & 2.04  \\
    MonoPair \cite{chen2020monopair} & 10.02 & 6.68 & 5.53 & 3.79 & 2.12 & 1.83  \\
    MonoRUn \cite{Chen2021MonoRUnM3} & 10.88 & 6.78 & 5.83 & 1.01 & 0.61 & 0.48  \\
    \midrule
    ImVoxelNet(+homo) & \textbf{12.47} & 7.62 & 6.72 & 1.52 & 0.85 & 0.94  \\
    MonoFlex(+homo) & 11.87 & \textbf{7.66} & \textbf{6.82} & 5.48 & 3.50 & 2.99  \\
    \bottomrule
    \end{tabular}}
    \end{center}
    \label{table:pedcyc}
 \end{table}

\textbf{Dataset and Evaluation Metrics.} Our proposed method is evaluated on KITTI 3D Object Detection benchmark ~\cite{Geiger2012CVPR}, which includes 7481 images for training and 7518 images for testing. The training set is split into 3712 samples for training and 3769 samples for validation as suggested in~\cite{NIPS2015_6da37dd3}. The classes are Car, Pedestrian, and Cyclist with three difficulty levels for each class, i.e., Easy, Moderate, and Hard. The official KITTI leaderboard is ranked on Moderate difficulty. Our method is evaluated on KITTI test set by submitting the detection results to the official server. For a fair comparison with other methods, we use official metrics, average precision (AP) with an IoU threshold of 0.7 for Car and 0.5 for both Pedestrian and Cyclist. In all experiments, the $\textup{AP}_{3D|R_{40}}$ results are reported for a comprehensive comparison with previous studies.

\textbf{Implementation Details.} We use the official implementations of ImVoxelNet\cite{rukhovich2021imvoxelnet} with ResNet50\cite{He2016DeepRL} and MonoFlex\cite{zhang2021objects} with DLA34\cite{Yu2018DeepLA} as their backbones. We follow all the experimental settings of the original code and add our homography loss as an auxiliary loss. For ImVoxelNet\cite{rukhovich2021imvoxelnet}, we add the loss at the beginning and train 24 epochs. As for MonoFlex\cite{zhang2021objects}, the homography loss is added after 40 epochs and we train the network 80 epochs in total. We name these two new implementations as \textbf{
ImVoxelNet(+homo)} and \textbf{MonoFlex(+homo)}, respectively.

\subsection{Quantitative Results}\label{subsec:results} 

\textbf{Results of Car category on KITTI \textit{test} set.} As demonstrated in Tab.~\ref{table:kittitest}, the proposed method MonoFlex(+homo) achieves superior results on Car category compared with the previous methods, even including those with extra data, such as depth or LiDAR point clouds. To be specific, MonoFlex(+homo) achieves 1.81\%, 1.05\%, and 1.00\% gains on the easy, moderate and hard settings, respectively. 
%For those methods with extra data like CaDDN\cite{reading2021categorical} and DFR-Net\cite{Zou_2021_ICCV}, our method also obtains about 1\%$\sim$2\% improvement in all different settings.
Besides, our ImVoxelNet(+homo) achieves 2.95\%, 2.02\%, and 1.45\% gains over the original baseline, which shows its robustness and effectiveness.

\textbf{Results of Car category on KITTI \textit{validation} set.} We also present our model's performance on the KITTI \textit{validation} set in Tab.~\ref{table:kittival}. Specifically, our method achieves the SOTA performance compared with the previous methods. 
%Compared to MonoPair~\cite{chen2020monopair} which also considers the mutual relationships, 
Compared to MonoPair~\cite{chen2020monopair}, our ImVoxelNet(+homo) and MonoFlex(+homo) get performance gain by 2.58\%/4.59\% for moderate setting at the 0.7 IoU threshold. This shows that our method is more capable of detecting hard examples in autonomous driving scenes by adding homography loss as an additional constraint.

\textbf{Pedestrian/Cyclist detection on KITTI \textit{test} set.} For Pedestrian and Cyclist, we present the detection performance in Tab.~\ref{table:pedcyc}. Our method MonoFlex(+homo) leads to the competitive performance in both categories. This shows our homography loss can also improve the performance for detecting small objects, e.g., human. MonoFlex(+homo) outperforms all other approaches in the Pedestrian category, with an 0.88\% improvement from the previous best method (7.66\% vs 6.78\%). A possible reason is that human's standing point is a more reliable reference for computing the homography matrix. % and our loss take the full advantage of the strong relationship between standing point and homography priority.

\subsection{Ablation Study}\label{subsec:ablation}

% We conduct ablation studies of our loss on Car category of the KITTI with default evaluation metric as $\textup{AP}_{3D|R_{40}}$. 

We conduct ablation studies to analyze the effects of our loss on Car category of the KITTI \textit{validation} set. The default evaluation metric is $\textup{AP}_{3D|R_{40}}$.

% \begin{table}[t]
%     \caption{Calculating homography with different configurations.}
%     \begin{center}
%     \scalebox{0.8}{
%     \begin{tabular}{c|c|c|c|c}
%     \toprule
%     \multirow{2}{*}{Method} & \multirow{2}{*}{Type} & \multicolumn{3}{|c}{$\textup{AP}_{3D|R_{40}}$}  \\ 
%     \cmidrule{3-5}
%      & & Easy & Moderate & Hard   \\
%     \midrule
%     \multirow{3}{*}{ImVoxelNet(+homo)} & 1 & 21.35 & 14.63 & 11.60  \\
%      & 2 & 21.44 & 14.88 & 12.08   \\
%      & Ensemble & 21.48 & 15.01 & 12.06   \\
%     \midrule
%     \multirow{3}{*}{MonoFlex(+homo)} & 1 & 23.04 & 16.89 & 14.90  \\
%      & 2 & 22.37 & 16.48 & 14.41  \\
%      & Ensemble & 22.93 & 16.80 & 14.55   \\
%     \bottomrule
%     \end{tabular}}
%     \end{center}
%     \label{table:calhomo}
%  \end{table}

\begin{table}[t]
    \caption{Different settings of ImVoxelNet are evaluated on the \textit{validation} set. By default, as shown in the top row, we combine homography type 2 (Sec.~\ref{subsubsec:calhomo}), representative proposal type 1 (Sec.~\ref{subsubsec:repreproposal}), and weight of 0.2 (Sec.~\ref{subsubsec:lossweight}) with homography loss to obtain the best performance. Each row evaluates one specific setting compared with the default choice. The bottom row shows the comparison with projection loss (Sec.~\ref{subsec:diff}).}
    \label{table:imvoxelnet}
    \begin{center}
    \scalebox{0.5}{
    \begin{tabular}{cc|ccc|ccccc|cc|ccc}
    \toprule
    \multicolumn{2}{c|}{Homo} & \multicolumn{3}{c|}{Proposal} & \multicolumn{5}{c|}{Weight} & \multicolumn{2}{c|}{Loss} & \multicolumn{3}{c}{$\textup{AP}_{3D|R_{40}}$} \\ \midrule
    1 & 2 & 1 & 2 & 3 & None & 0.1 & 0.2 & 0.5 & 1.0 & +homo & +proj & Easy & Moderate & Hard \\ \midrule
     & \CheckmarkBold & \CheckmarkBold &  &  &  &  & \CheckmarkBold &  &  & \CheckmarkBold &  & \textbf{21.44} & \textbf{14.88} & \textbf{12.08} \\ \midrule
    \checkmark &  & \checkmark &  &  &  &  & \checkmark &  &  & \checkmark &  & 21.35 & 14.63 & 11.60 \\ \midrule
     & \checkmark &  & \checkmark &  &  &  & \checkmark &  &  & \checkmark &  & 19.41 & 14.21 & 11.63 \\
     & \checkmark &  &  & \checkmark &  &  & \checkmark &  &  & \checkmark &  & 20.29 & 14.26 & 11.60 \\ \midrule
     & \checkmark & \checkmark &  &  & \checkmark &  &  &  &  & \checkmark &  & 20.20 & 13.85 & 11.41 \\
     & \checkmark & \checkmark &  &  &  & \checkmark &  &  &  & \checkmark &  & 21.01 & 14.19 & 11.53 \\
     & \checkmark & \checkmark &  &  &  &  &  & \checkmark &  & \checkmark &  & 20.43 & 14.13 & 11.48 \\
     & \checkmark & \checkmark &  &  &  &  &  &  & \checkmark & \checkmark &  & 19.27 & 13.99 & 11.53 \\ \midrule
     &  &  &  &  &  &  & \checkmark &  &  &  & \checkmark & 20.51 & 14.13 & 11.49 \\ \bottomrule
    \end{tabular}
    }
    \end{center}
\end{table}

\subsubsection{Calculating Homography}\label{subsubsec:calhomo}

To calculate the homography matrix, we use $\mathbf{q}_{gt}$ and $\tilde{\mathbf{Q}}_{pred}$ (Type 1) to construct the geometric constraints. Similarly, $\mathbf{q}_{pred}$ and $\tilde{\mathbf{Q}}_{gt}$ (Type 2) can also be selected. Therefore, we compare the performance of these two types in ImVoxelNet(+homo) and MonoFlex(+homo). The results are listed in Tab.~\ref{table:imvoxelnet} and~\ref{table:monoflex}. We can see that for those methods that predict in BEV domain like ImVoxelNet, Type 2 is more suitable. As for those who predict in 2D images like MonoFlex, Type 1 gets higher performance. Therefore, the prediction domain can affect the final performance. So how to choose a proper type will finally depend on the specific application.

% \begin{table}[t]
%     \caption{Different settings of representative proposal evaluated on ImVoxelNet(+homo).}
%     \label{table:repreproposal}
%     \begin{center}
%     \scalebox{0.8}{
%     \begin{tabular}{@{}c|ccc@{}}
%     \toprule
%     \multirow{2}{*}{Type} & \multicolumn{3}{c}{$\textup{AP}_{3D|R_{40}}$}                                           \\ \cmidrule(l){2-4} 
%                           & \multicolumn{1}{c|}{Easy}  & \multicolumn{1}{c|}{Moderate} & Hard  \\ \midrule
%     1                     & \multicolumn{1}{c|}{21.44} & \multicolumn{1}{c|}{14.88}    & 12.08 \\
%     2                     & \multicolumn{1}{c|}{19.41} & \multicolumn{1}{c|}{14.21}    & 11.63 \\
%     3                     & \multicolumn{1}{c|}{20.29} & \multicolumn{1}{c|}{14.26}    & 11.60 \\ \bottomrule
%     \end{tabular}}
%     \end{center}
% \end{table}

\subsubsection{Representative Proposals}\label{subsubsec:repreproposal} In anchor-based methods like ImVoxelNet, several anchors will be assigned to the same ground truth box based on IoU threshold. Therefore, we need to select the representative proposal from these positive proposals. Here, we have three strategies of selection: 1) the proposal with the highest classification score, 2) the proposal with the highest IoU score, 3) the average proposal of all positive anchors. We conduct the ablation experiment in Tab.~\ref{table:imvoxelnet}. The result shows that the one with the highest classification score achieves the best performance at 14.88\% of the moderate setting. It also shows our homography loss can strengthen the consistency between regression and classification heads.

\subsubsection{Replicated Losses}\label{subsubsec:repliproposal}
% \jq{the title may need to change Replicated Boxes or Replicated Losses}
For anchor-free methods, such as MonoFlex, the depth regression head can be very unstable at the beginning of training. To solve this problem, we refer to the replicated strategy in \cite{park2021pseudo} and propose a replicated proposal strategy here to strengthen the robustness. The homography loss is replicated 3 times in total to get a reliable homography matrix. We conduct the ablation by four different settings: 1) $\mathbf{q}_{pred}$ + $\mathbf{Depth}_{pred}$ (the predicted depth), 2) $\mathbf{q}_{pred}$ + $\mathbf{Depth}_{gt}$ (the ground truth depth). 3) $\mathbf{q}_{gt}$ + $\mathbf{Depth}_{pred}$. 4) ensemble by adding the aforementioned three losses together. The results are shown in Tab.~\ref{table:monoflex}. We observe that the ensemble strategy has a better result due to sufficient constraints.

\subsubsection{Loss Weight}\label{subsubsec:lossweight}

To determine the final loss weight of ImVoxelNet(+homo) and MonoFlex(+homo), we also conduct experiments on loss weights. The results are shown in Tab.~\ref{table:imvoxelnet}. We observe superior performance when the loss weight is 0.2. Therefore, we apply this configuration in training and get the final performance. For MonoFlex(+homo), we also do the same experiments and get 0.2 as a result. It shows our homography loss can be served as an auxiliary loss for detection.

\begin{table}[t]
    \caption{Different settings of MonoFlex are evaluated on the \textit{validation} set. By default, as shown in the top row, we combine homography type 1 (Sec.~\ref{subsubsec:calhomo}), ensembled losses (Sec.~\ref{subsubsec:repliproposal}), and weight of 0.2 with homography loss to achieve the best result.}
    \label{table:monoflex}
    \begin{center}
    \scalebox{0.6}{
    \begin{tabular}{cc|cccc|c|c|ccc}
    \toprule
    \multicolumn{2}{c|}{Homo} & \multicolumn{4}{c|}{Replicated losses} & Weight & Loss & \multicolumn{3}{c}{$\textup{AP}_{3D|R_{40}}$} \\ \midrule
    1 & 2 & 1 & 2 & 3 & Ensemble & 0.2 & +homo & Easy & Moderate & Hard \\ \midrule
    \CheckmarkBold &  &  &  &  & \CheckmarkBold & \CheckmarkBold & \CheckmarkBold & \textbf{23.04} & \textbf{16.89} & \textbf{14.90} \\ \midrule
     & \checkmark &  &  &  & \checkmark & \checkmark & \checkmark & 22.37 & 16.48 & 14.41 \\ \midrule
    \checkmark &  & \checkmark &  &  &  & \checkmark & \checkmark & 21.92 & 16.54 & 13.84 \\
    \checkmark &  &  & \checkmark &  &  & \checkmark & \checkmark & 22.48 & 16.62 & 14.49 \\
    \checkmark &  &  &  & \checkmark &  & \checkmark & \checkmark & 22.51 & 16.69 & 14.46 \\ \bottomrule
    \end{tabular}
    }
    \end{center}
\end{table}

\begin{figure*}[t]
    \centering
    \includegraphics[width=0.83\linewidth]{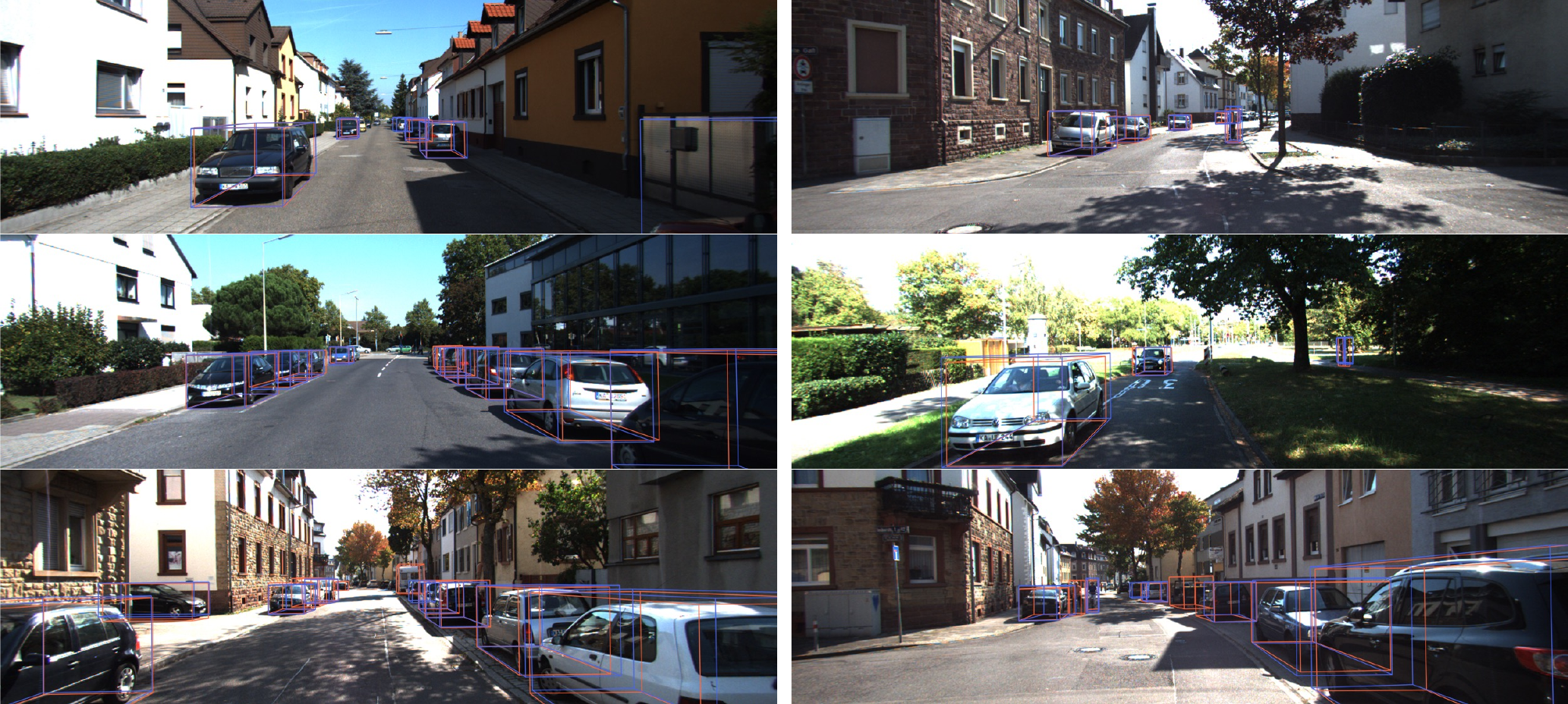}
    \caption{We visualize the results of 3D object detection using ImVoxelNet(+homo) on KITTI \textit{val} set, where the orange represents the ground truth and our predicted results are colored in blue. The left column shows results of the network trained on the Car category only, and the right column is trained on three categories including Car, Pedestrian, and Cyclist. It is worth noting that, with the homography loss, it is possible to detect small targets and even truncated objects.}
    \label{fig:5}
\end{figure*}

\subsection{Qualitative Results}

From the qualitative results demonstrated in Fig.~\ref{fig:5}, with the proposed homography loss function, we can get superior performance for normal objects in the scene. Even for very challenging cases, such as small objects (distant Pedestrian and Car), and extremely truncated objects, our method can still successfully detect those well.% which shows the effectiveness of our proposed method.

% \bj{\subsection{Extension to LiDAR based method}}

\section{Discussions}\label{sec:discussions}

\subsection{Difference with Projection Loss}\label{subsec:diff}

% \begin{table}[t]
%     \caption{TBD}
%     \label{tab:homo_cmp_proj}
%     \begin{center}
%     \scalebox{0.9}{
%     \begin{tabular}{@{}c|ccc@{}}
% \toprule
% \multirow{2}{*}{ImVoxelNet} & \multicolumn{3}{c}{$\textup{AP}_{3D|R_{40}}$}                                           \\ \cmidrule(l){2-4} 
%                         & \multicolumn{1}{c|}{Easy} & \multicolumn{1}{c|}{Moderate} & Hard \\ \midrule
% +homo       & \multicolumn{1}{c|}{\textbf{21.44}}     & \multicolumn{1}{c|}{\textbf{14.88}}         &   \textbf{12.08}  \\
% +proj       & \multicolumn{1}{c|}{20.51}     & \multicolumn{1}{c|}{14.13}         &   11.49  \\ \bottomrule
% \end{tabular}
% }
%     \end{center}
% \end{table}

As shown in Eq.~\ref{eq:proj}, the calibration parameters of the camera can be used to project a single predicted 3D keypoint onto a 2D image plane which will be further constrained by its corresponding 2D ground-truth value. It means that each training sample is considered individually, and the predicted 3D positions are also refined and optimized independently during network training. This is the key idea of the commonly used projection loss. However, for calculating the homography matrix, all pairs of correspondences will be involved in the computation, each pair of corresponding 2D/3D points will contribute two linear equations for solving Eq.~\ref{eq:calc_H}. During backpropagation of the gradient of Eq.~\ref{eq:loss_homo}, $\hat{\mathbf{H}}$ is gradually optimized, that is to say, all the predicted $\tilde{\mathbf{Q}}_{pred}$ that are used for calculating the homography matrix will also be refined according to the chain rule. Therefore, homography loss can be leveraged to globally constrain the optimization of 3D localization. We compare projection loss with the proposed homography loss as shown in Tab.~\ref{table:imvoxelnet}.

\subsection{Depth Range Statistics}

\begin{table}[t]
    \caption{Depth range statistics at the 0.7 and 0.5 IoU threshold.}
    \label{tab:range_test}
    \begin{center}
    \scalebox{0.7}{
    \begin{tabular}{@{}c|c|cccc@{}}
    \toprule
    \multirow{2}{*}{Metric} & \multirow{2}{*}{ImVoxeNet} & \multicolumn{4}{c}{Depth Range (m)} \\ \cmidrule(l){3-6} 
     &  & 0-10 & 10-20 & 20-30 & \textgreater{}30 \\ \midrule
    \multirow{4}{*}{\begin{tabular}[c]{@{}c@{}}KITTI\\ Moderate \\ $\textup{AP}_{3D|R_{40}}$ \end{tabular}} & baseline@0.7 & 35.45 & 17.48 & 1.23 & 0.17 \\
     & +homo@0.7 & \textbf{35.99} & \textbf{20.48} & \textbf{2.17} & \textbf{0.20} \\
     & baseline@0.5 & 78.57 & 59.34 & 15.24 & 3.31 \\
     & +homo@0.5 & \textbf{81.57} & \textbf{61.68} & \textbf{18.44} & \textbf{4.07} \\ \bottomrule
    \end{tabular}}
    \end{center}
\end{table}

In order to investigate why homography loss is useful for improving the accuracy of 3D detection. We design an experiment that divides the depth range into several segments as shown in Tab.~\ref{tab:range_test} and gets the statistics for each interval. For fairness, the evaluation metric is also $\textup{AP}_{3D|R_{40}}$ on the Car category of the KITTI \textit{validation} set with the difficulty of moderate level. Obviously, we can see that in the area of 10 meters away, the effect of the detection algorithm with homography loss is much better than that of the baseline. Especially in the range of 10-20m, we obtain 3.0\% and 2.34\% gains over the baseline method with different IoU thresholds, respectively. This shows that our loss function is more effective for small target detection. The reason is that, as elaborated in Eq.~\ref{eq:calc_H}, the ground truth 2D position $\mathbf{q}_{gt}$ on the image plane is used as guidance to correct the predicted 3D position $\tilde{\mathbf{Q}}_{pred}$. 
% \bj{The relative geometric relationship of objects on image plane will also be preserved after homography transformation, it implied that the relative 3D positions (predicted) on BEV plane also needs to be consistent with the 2D counterparts, which is explicitly enabled by homography loss.}
The relative geometric relationship of objects on the image plane will be transferred to the corresponding 3D objects on BEV plane by homography loss.
% \bj{In other words, as shown in Tab.~\ref{tab:range_test}, the performance of near object detection has not been improved very obviously, and the accuracy itself is already high, but the improvement of the detection effect of distant objects is due to the homography relationship, which further refines the inaccurate estimated 3D positions in order to satisfy the overall geometric constraint.}
In other words, the improvement of the detection effect of distant objects in Tab.~\ref{tab:range_test} is due to the homography relationship, which refines the inaccurate estimated 3D positions to satisfy the overall geometric constraint.

\subsection{Limitations}

As stated in Sec.~\ref{subsec:homo_loss}, we assume that the ground plane is flat and use the simplified 2D coordinates $\hat{\mathbf{Q}}=[x,y]^T$ on BEV plane to replace the original 3D points $\mathbf{Q} = [x,y,z]^T$. However, in practice, as pointed out in \cite{Zhou_2021_CVPR}, usually the road is not smooth and has slight fluctuation, it will influence the accuracy of 3D detection. 
% This will be our future work to explore by possibly introducing the uncertainty estimation of the predicted 3D positions.

\section{Conclusion}

In this paper, we propose a differentiable loss function, named as \textit{homography loss}, which is a plug-and-play module that can be integrated into any monocular 3D detector, to help globally optimize the 3D positions of all the objects, instead of taking each object as an independent sample during training. Homography loss also fully exploits the inherent connection between 2D image space and 3D Bird's Eye View and constrains the optimization of 3D positions under the guidance of 2D localization, which is demonstrated to be useful for detecting small targets or highly truncated objects. In the future work, we will consider how to avoid the assumption of flatness of the ground. 

% However, our assumption is that the ground plane should be flat, whereas usually the road is not smooth and has slight fluctuation, which can influence the accuracy of 3D detection. This will be our future work.

%However, as stated in Sec.~\ref{subsec:homo_loss}, we use the simplified 2D coordinates $\hat{\mathbf{Q}}=[x,y]^T$ on BEV plane to replace the original 3D bottom points $\mathbf{Q} = [x,y,z]^T$, it is because the premise of this assumption is that the ground plane is flat. However, in practice, as pointed out in \cite{Zhou_2021_CVPR}, usually the road is not smooth and has slight fluctuation, it can also influence the accuracy of 3D detection. This will be our future work. % to explore by possibly introducing the uncertainty estimation of the predicted 3D positions.

% \section{Acknowledgement}

\paragraph{Acknowledgments:} This work was (partially) supported by the National Key R\&D Program of China under Grant 2020AAA0103902, NSFC-Zhejiang Joint Fund for the Integration
of Industrialization and Informatization under Grant U1709214 and Key Research \& Development Plan of Zhejiang Province (2021C01196).

%%%%%%%%% REFERENCES
{\small
\bibliographystyle{ieee_fullname}
\bibliography{egbib}
}

\end{document}